\documentclass{article}

\usepackage{arxiv}

\usepackage[utf8]{inputenc} 
\usepackage[T1]{fontenc}    
\usepackage{hyperref}       
\usepackage{url}            
\usepackage{booktabs}       
\usepackage{amsfonts}       
\usepackage{nicefrac}       
\usepackage{microtype}      
\usepackage{lipsum}		
\usepackage{graphicx}
\usepackage{natbib}
\usepackage{doi}

\title{Quick Starting Dialog Systems with Paraphrase Generation}

\author{{Louis Marceau, Raouf Belbahar, Marc Queudot}\\ 
	National Bank of Canada\\
	Montreal, QC, Canada\\
	\texttt{\{louis.marceau,raoufmoncef.belbahar,marc.queudot\}@bnc.ca}\\
    \And{Nada Naji}\\
    \texttt{nada.aj.naji@gmail.com} \\
    \AND {Éric Charton}\\
	National Bank of Canada\\
	Montreal, QC, Canada\\
	\texttt{eric.charton@bnc.ca} \\
	\And
	\href{https://orcid.org/0000-0001-8196-2153}{\includegraphics[scale=0.06]{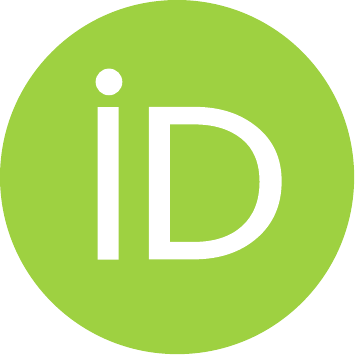}\hspace{1mm}Marie-Jean Meurs} \\
	Université du Québec à Montréal\\
	Montreal, QC, Canada\\
	\texttt{meurs.marie-jean@uqam.ca} \\
}

\date{}

\renewcommand{\shorttitle}{Quick Starting Dialog Systems with Paraphrase Generation}

\hypersetup{
pdftitle={Quick Starting Dialog Systems with Paraphrase Generation},
pdfsubject={cs.CL},
pdfauthor={Louis Marceau, Raouf Belbahar, Marc Queudot, Eric Charton, Marie-Jean Meurs},
pdfkeywords={Chatbot, Dialog Systems, Paraphrase Generation, Natural Language Processing},
}

\begin{document}
\maketitle

\begin{abstract}
Acquiring training data to improve the robustness of dialog systems can be a painstakingly long process. In this work, we propose a method to reduce the cost and effort of creating new conversational agents by artificially generating more data from existing examples, using paraphrase generation. 
Our proposed approach can kick-start a dialog system with little human effort, and brings its performance to a level satisfactory enough for allowing actual interactions with real end-users. 
We experimented with two neural paraphrasing approaches, namely Neural Machine Translation and a Transformer-based seq2seq model. 
We present the results obtained with two datasets in English and in French:~a crowd-sourced public intent classification dataset and our own corporate dialog system dataset. 
We show that our proposed approach increased the generalization capabilities of the intent classification model  on both datasets, reducing the effort required to initialize a new dialog system and helping to deploy this technology at scale within an organization.
\end{abstract}


\section{Introduction}

Dialog systems are becoming increasingly common as personal assistants and information seeking interfaces. 
In the context of customer support, dialog systems are nowadays playing an important role as an additional communication channel to respond to recurrent client queries~\cite{ganhotra2019knowledgeincorporating}. 
However, insufficient volume of available training data as well as the huge effort required to gather such data hinder the development of robust domain-specific conversational AI systems~\cite{schuster2019crosslingual}. 
Dialog systems go through a long process towards maturity where considerable volumes of data need to be gathered to reach satisfactory performance levels, suitable for real users. 
In the context of question-answering dialog systems, we usually refer to intents as the categorization of the various information needs or possible questions that users have. 
These intents are defined using existing observations, and they act as classes in the Natural Language Understanding (NLU) model of the dialog system. 
Such a model attempts at associating an incoming user message to one of the predefined intents to subsequently select a relevant response. 

Users can express the same question in different ways. 
For instance, \texttt{"I forgot my password, what can I do?"} and \texttt{"don't remember my pass-code"} are both associated  with the same intent but are phrased differently. 
Creating large enough volumes of training examples that cover such variations is a labor-intensive process. 
This is especially true for dialog systems that contain hundreds of intents. 
When building a new dialog system in a real industry setting, the initial dataset of observations is often constructed from Frequently Asked Questions (FAQ) pages or questions asked to human agents. 
While this dataset should be as rich and complete as possible to be used as a training set, it is rarely possible to collect multiple samples per intent at this point. 
A certain amount of training samples per intent must hence be acquired in order to reach sufficient performance level for the system to be exposed to real users with minimal risks.
In addition to that, since different dialog systems cover different use-cases, this data acquisition task must be repeated every time a new system is created. 
This can be problematic when trying to deploy these systems in an organization at scale. 
As the number of dialog systems increases, so do the resources and effort required to generate and maintain them.

The solution we propose to this problem relies on generating artificial paraphrases based on existing observations during the design of a new dialog system, in order to increase the volume of training samples per intent.
The artificially generated paraphrases will help increase the performance of the intent classification model

Before deploying a new dialog system in an industrial context, the system must reach a certain level of performance - above a given threshold - in order to minimize the risk and dissatisfaction associated with a poorly-functioning system. 
Reaching this threshold usually comes by acquiring more data per intent. 
Artificial samples can help increase the intent classification model generalization capabilities above this threshold, hence allow the system to be exposed to end-users.

In our context, training samples for a given intent are paraphrases because they use different words and syntax but have the same semantic content. 
The goal of this task is not to replace real data, but rather to bootstrap the dialog system while avoiding the cost of manually generating more data.

To achieve the paraphrase generation task, we compare different neural approaches in order to augment our training data. 
The first is based on Neural Machine Translation (NMT)~\cite{Mallinson2017Paraph}. 
In this approach, data samples are translated into another language, referred to as the \emph{pivot language} and then the result is translated back to the source language.
The translations obtained after this step will be considered as potential paraphrases.

The second method we propose consists in directly outputting paraphrases using a neural network trained for text generation. 
This neural network uses the encoder-decoder architecture of the Transformer~\cite{vaswani2017attention}, which has achieved state-of-the-art results in many Natural Language Processing (NLP) tasks. 
Large transformer-based language models (LMs) like GPT-2~\cite{Radford2019GPT-2}, BERT~\cite{devlin2019bert} or T5~\cite{raffel2020exploring} can learn powerful representations of the data and can also be fine-tuned for a specific downstream task, like paraphrase generation.

Our approach significantly improves the performance of the intent classification model with little to no human intervention, hence, reduces both the cost to build an initial training set and the time-to-market to deploy the dialog system. Once the system is put into production it can start gathering real data from interactions with the end users. Using this approach, dialog systems can be deployed at scale in an organization without the huge cost associated with handcrafting training data for every intent.

The rest of this paper is organized as follows: Section~\ref{sec:review} surveys related previous work in the domain of dialog systems, dataset availability and augmentation. 
Section~\ref{sec:data} presents in details the datasets that we used in our experiments. In Section~\ref{sec:design}, we lay out the experimental setup of our work including the architecture of the proposed approaches. 
The evaluation framework is presented in Section~\ref{sec:eval}. 
Section~\ref{sec:results} reports the results achieved on the various intent classification datasets and discusses the implication of using paraphrasing in training set augmentation for NLU models. 
Finally, Section~\ref{sec:conclusion} concludes this work, and explores possible future horizons.

\section{Related Work} \label{sec:review}
Dialog systems have evolved tremendously in the last decade following advances in deep learning. 
Question answering systems and virtual assistants such as Alexa and Siri have become more and more widespread~\cite{Gao2019NeuralApp,Chen2017Deep,Chen2017Survey}.
When it comes to building new dialog systems, one of the main challenges is obtaining conversational training data. 
Several initiatives have made available conversational datasets such as the Ubuntu Dialog Corpus \cite{Lowe2015Ubuntu}, the crowd-sourced, publicly available intent classification dataset~\cite{Kang2018DataCF}, the in-scope and out-of-scope intents dataset CLINC150~\cite{Larson2019AnED}, and the CAsT-19 Dataset (TREC Complex Answer Track)~\cite{Dalton2020Dataset}. 
\cite{serban2015survey} surveyed publicly available datasets for Dialog systems development. 
The authors often describe the process of building such datasets as, unsurprisingly, expensive and labor-intensive. 
While such datasets constitute a valuable source for exploration, they have limited use when it comes to building production-quality, domain-specific systems, for instance in the medical or financial domains\footnote{While there exists several types of dialog systems, our present work focuses solely on question answering dialog systems.}.

Data augmentation is often used in computer vision and image processing. 
Basic transformations such as cropping, rotating, and addition of Gaussian noise are applied to images to increase the size of the training data and reduce overfitting~\cite{Badmimala2019Augment,Krizhevsky2012CompVision,Deng2009CompVision}. 
NLP tasks, such as text classification or detection of sentence similarity, leverage this idea by applying transformations like shuffling (changing the order of tokens or sentences), or incorporating synonyms and embeddings~\cite{Badmimala2019Augment,Wang2015Annoy,Papadaki2017DataAT}. 
Alternatively, paraphrasing can be used to transform the text by conveying the same meaning in alternative ways in the same natural language. 
Several NLP methods have been used in the past to generate paraphrases. 
Rule-based approaches were explored with handcrafted sets of rules or patterns to generate paraphrases~\cite{Mckeown1979Paraphrasing,Meteer1988Paraphrasing}. 
Other methods use monolingual or bilingual corpora approaches.
For instance,~\cite{Bannard2005Paraphrasing} proposed an approach using a bilingual corpus with pairs of translated sentences to obtain paraphrases in English by pivoting through another language, such as French or German. 
This idea was pushed even further by using NMT models to pivot, and generate paraphrases automatically in the source language~\cite{Mallinson2017Paraph,Sokolov2020NeuralMT}. 
In this direction,~\cite{Junczys2018Marian} introduced Marian, an open-source NMT framework developed on a deep-learning back-end based on reverse-mode auto-differentiation with dynamic computation graphs. 
We describe our use of Marian in Section~\ref{subsec:nmt-system}. 
More recently, large language models have shown to be extremely effective for many NLP tasks including text generation. Such models have been fine-tuned with a paraphrasing dataset to serve as paraphrasing generators~\cite{Witteveen2019Paraphrasing}.
Paraphrasing has been used in summarization and plagiarism detection tasks~\cite{10.1162/COLI_a_00153}, among others, but has not been thoroughly explored in augmenting conversational corpora, to the best of our knowledge. 
In this paper, we propose a low-cost approach to rapidly grow a small training dataset with the use of paraphrasing based on machine translation and large language models. The following Section details the datasets used to test our data augmentation methods.

\section{Datasets} \label{sec:data}

We used two different datasets in our experiments. 
The first dataset is the CLINC150 dataset \footnote{\url{https://github.com/clinc/oos-eval}}. 
It was built for the task of evaluating intent classification~\cite{Larson2019AnED}. 
This dataset contains 150 intents and some out-of-scope data, which means samples that are not associated with one of the intents. 
The intents mainly correspond to task-oriented queries (e.g., "move 100 dollars from my savings to my checking") though some queries related to general information can be found as well (e.g., "how do I use my vacation days"). 
Out-of-scope data constitute utterances or messages that are not associated with any in-scope intents and hence do not carry any interest for the task at hand. 
Therefore, we only used the in-scope dataset in our experiments. 

The in-scope dataset (SCOPE) is composed of 22,500 queries covering 150 intents, which correspond to ten domain : \emph{banking}, \emph{work}, \emph{meta}, \emph{Auto \& commute}, \emph{home}, \emph{travel}, \emph{utility}, \emph{kitchen and dining}, \emph{small talk} and \emph{credit cards}. 
SCOPE is split into  training, validation and testing sets. 
The training set contains 100 samples per intent, thus yielding 15,000 training samples in total, while the validation and test have 20 and 30 samples per intent respectively. 
We made several partitions of the training set by randomly sampling 5, 10 and 50 training samples per intent in order to measure the impact of paraphrasing on datasets of various sizes. We refer to these sets as \emph{SCOPE-train-5}, \emph{SCOPE-train-10}, and \emph{SCOPE-train-50}, respectively.
The original testing set was used to evaluate the different data augmentation methods. 
It is composed of 4,500 training samples, 30 samples for each of the 150 intent. We refer to this set as \emph{SCOPE-test}.
Finally, we left out the validation set since the configuration and hyperparameters of the NLU model were kept constant throughout the various experiments.

The second dataset is an in-house dialog dataset, which originates from the first iteration of a question answering dialog system currently deployed on the corporate website and transactional platform. We refer to this dataset as HOUSE. 
It was built by defining intents based on questions extracted from a FAQ Section of the corporate website, and manually creating some different formulations for each. 
This was done as a first step towards developing a dialog system to answer clients' frequent questions, such as how to change their password or perform a specific transaction. 
 
We then split this handcrafted dataset into training and testing sets. 
It is worth mentioning that the HOUSE dataset is entirely in French. 
The training set consists of 73 training samples, each of the 20 intents is associated to three to six samples.
We refer to this set as \emph{HOUSE-train}. 
The test set is composed of 77 samples in total, with one to 11 samples per intent. 
We call this set the \emph{HOUSE-test}. 
We used a more recent version of the test set in order to have more testing samples, but only the starting 20 intents were kept. This is why it contains more samples than the training set.
 
Tables~\ref{tab:datasets_simple} and \ref{tab:datasets_all} summarize the characteristics of aforementioned datasets.
 
\begin{table*}[t]
\centering
{\def\arraystretch{1.2}
\begin{tabular}{|l|c|c|c|l|}
\hline
\fontsize{7pt}{8pt}\selectfont\textbf{Dataset} & \fontsize{7pt}{8pt}\selectfont\textbf{Language} & \textbf{\begin{tabular}[c]{@{}c@{}}\fontsize{7pt}{8pt}\selectfont Total \\\fontsize{7pt}{8pt}\selectfont  intents\end{tabular}} & \textbf{\begin{tabular}[c]{@{}c@{}}\fontsize{7pt}{8pt}\selectfont Total\\\fontsize{7pt}{8pt}\selectfont samples\end{tabular}} & \textbf{\begin{tabular}[c]{@{}c@{}}\fontsize{7pt}{8pt}\selectfont Examples\end{tabular}}\\
\hline
HOUSE & French & 20 & 150 &\begin{tabular}[l]{@{}l@{}}"Comment me connecter au site en ligne si j'ai \\déjà un compte" (translation: How can I login to \\the website if I already have an account)\end{tabular}\\
\hline
SCOPE & English & 150 & 5,250 - 12,000 & \begin{tabular}[l]{@{}l@{}}"Is rice ok after 3 days in the refrigerator" \\(kitchen and dining), "How can I see my rewards\\for my visa card" (banking)\end{tabular}\\
\hline
\end{tabular}}
\vspace{0.1in}
\caption{Datasets used in our experiments. HOUSE and SCOPE, respectively, represent our in-house corporate dataset and the in-scope public dataset. Their respective source language and sizes are given, along with an example of query.}
\label{tab:datasets_simple}
\end{table*}

\section{Methodology and Experiments} \label{sec:design}

In this Section, we describe our proposed approaches for dataset augmentation. Sub-section~\ref{subsec:nmt-system} presents the NMT-based paraphrasing, pivoting around different languages such as English, German and Spanish for the HOUSE dataset (which is in French), and French and German for the SCOPE dataset. 
Sub-section~\ref{subsec:language-model-system} presents the second approach, which utilizes a fine-tuned PEGASUS~\cite{zhang2020pegasus} language model for the task of paraphrase generation.

\subsection{The NMT-based System} \label{subsec:nmt-system}

The pre-trained NMT models we used are composed of Transformer models~\cite{vaswani2017attention} along with guided word alignments based on the Marian-NMT framework~\cite{Junczys2018Marian}. 
The required pre-processing for these models consists of normalizing punctuation and using the language-independent subword tokenizer SentencePiece~\cite{kudo2018sentencepiece} to get a subword-based tokenization while benefiting from the regularization algorithm introduced in the paper. 
The subword tokenizer offers the best of both word-based and character-based tokenization:~known words are taken into account in the vocabulary, and the tokenizer can also make use of word pieces and individual characters for unknown words.

We use a two-step process, each step involving a NMT model. 
During the first step, training samples are passed through the source-to-pivot (e.g. English-to-German) model and we keep the best translation for each sample (ranked by the NMT model). 

In the second step, we translate back to the source language (e.g. German-to-English) to generate several variations. Often the first best translation (as ranked by the NMT model) is identical to the original sample, and is therefore systematically removed. 
At this point, we keep the following $n$-best translations as paraphrases of the original input test. 
We experimented with varying the value of $n$, and found that setting $n$ to six provides relevant syntactical and lexical variations without causing a sharp semantic drift from the original samples. 

\subsection{The LM-based System} \label{subsec:language-model-system}

This system leverages a pre-trained transformer-based model using PEGASUS pre-training method. 
The PEGASUS self-supervised pre-training method is designed for abstractive summarization which transfer well for our paraphrase generation task. In PEGASUS, important sentences are removed/masked from an input document and the model is tasked with recovering and outputting them in a sequence, similar to an extractive summary ~\cite{zhang2020pegasus}. In the fashion of transformer model for text generation, it's been pre-trained on a large corpus of web-crawled documents.

For our experiments, we used a version of PEGASUS that had been fine-tuned for paraphrase generation\footnote{\url{https://huggingface.co/tuner007/pegasus_paraphrase}}. 
The dataset used for fine-tuning was filtered out from multiple paraphrasing datasets like Google's PAWS dataset~\cite{pawsx2019emnlp}.

When using this system on a dataset in French (HOUSE), we added a translation step using the Marian NMT framework. 
In this case, we use the top translation (as scored by the model) for each sample in the dataset as the input for the paraphrasing model. 
Once again, we take the 6-best (top scored by the model) generated paraphrases for each sample and translate them back to the source language. 

Once the paraphrases have been generated using either approach (NMT- or LM-based), the next step in the process is the detection and removal of duplicates. 
The goal of this step is to remove repetitive formulations. 
To perform this de-duplication, the original samples and the generated paraphrases are passed through a normalizer that performs case-folding (to lower-case) and de-punctuation. 
Afterwards, any paraphrase that matches any of the original samples associated to a given intent is deleted. The outcome of this step is a set of unique samples and paraphrases associated to each intent. 
This outcome constitutes the augmented training dataset.

As concrete examples, consider the sample \emph{"I must apply for a new credit card"}, which was translated by the NMT to \emph{"Ich muss eine neue Kreditkarte beantragen"} and \emph{"Je dois demander une nouvelle carte de crédit"} using German and French as pivot languages respectively. 
Below we report three paraphrases generated from the 6-best set: 
\\
With German as a pivot language, \emph{"I need to apply for a new credit card"}, \emph{"I have to request a new credit card"}, and \emph{" I need to request a new credit card"}; 
\\

With French as a pivot language: \emph{"I need to ask for a new credit card"}, \emph{"I need to request a new credit card"}, and \emph{"I need to apply for a new credit card"}. 
\\
As can be seen, there are some language-specific subtleties between the paraphrases, such as the use of \emph{"ask"} from the French pivot, which is not present in the former set wherein German was used for pivoting.
As for the PEGASUS-generated paraphrases, the following ones were generated for the same example: \emph{"I need a new credit card"}, \emph{"I need to apply for new credit card"}, and \emph{"I have to apply t for a new credit card."} and, interestingly, \emph{"I need a new credit card because I'm looking for a job"} which depicts the drift that natural language generation can sometimes introduce. It can be useful to introduce this additional information in the training data as long as it is something a user could say for this specific intent and it is not completely changing the meaning of the sentence.

\subsection{Baselines and Augmented Datasets} 

\noindent\textbf{The HOUSE dataset.} The core training set HOUSE-train constitutes the baseline for augmentation in this dataset. 
To evaluate our augmentation approach using NMT- and LM-based approaches, we generated the following datasets using the steps described in Sub-sections~\ref{subsec:nmt-system} and~\ref{subsec:language-model-system}: 
\begin{itemize}
    \item HOUSE-paraph (NMT-en): paraphrased with English as the pivot language to generate paraphrases in French for each intent. 
    \item HOUSE-paraph (NMT-de): paraphrased with German as the pivot language to generate paraphrases in French for each intent. 
    \item HOUSE-paraph (NMT-es): paraphrased with Spanish as the pivot language to generate paraphrases in French for each intent. 
    \item HOUSE-paraph (LM): augmented using PEGASUS (translation back and forth provided by the Marian NMT framework).
\end{itemize}

The HOUSE-train baseline and all the augmented datasets are in French.\\

\begin{table*}[t]
\centering
{\def\arraystretch{1.2}
\begin{tabular}{|l|r|r|r|r|r|}
\hline
\fontsize{7pt}{8pt}\selectfont\textbf{Dataset} & \textbf{\begin{tabular}[c]{@{}c@{}}\fontsize{7pt}{8pt}\selectfont Number \\\fontsize{7pt}{8pt}\selectfont of  intents\end{tabular}} & \textbf{\begin{tabular}[c]{@{}c@{}}\fontsize{7pt}{8pt}\selectfont Total\\\fontsize{7pt}{8pt}\selectfont samples\end{tabular}} & \textbf{\begin{tabular}[c]{@{}c@{}}\fontsize{7pt}{8pt}\selectfont Min samples\\\fontsize{7pt}{8pt}\selectfont per intent\end{tabular}} & \textbf{\begin{tabular}[c]{@{}c@{}}\fontsize{7pt}{8pt}\selectfont Max samples\\\fontsize{7pt}{8pt}\selectfont per intent\end{tabular}} & \textbf{\begin{tabular}[c]{@{}c@{}}\fontsize{7pt}{8pt}\selectfont Avg samples\\\fontsize{7pt}{8pt}\selectfont per intent\end{tabular}}\\
\hline
\fontsize{7pt}{8pt}\selectfont
HOUSE-train & 20 & 73 & 3 & 6 & 3.6 \\
\fontsize{7pt}{8pt}\selectfont
HOUSE-paraph (NMT-en) & 20 & 453 & 17 & 42 & 22.6 \\
\fontsize{7pt}{8pt}\selectfont
HOUSE-paraph (NMT-de) & 20 & 463 & 16 & 41 & 23.15 \\
\fontsize{7pt}{8pt}\selectfont
HOUSE-paraph (NMT-es) & 20 & 458 & 16 & 41 & 22.9 \\
\fontsize{7pt}{8pt}\selectfont
HOUSE-paraph (LM) & 20 & 351 & 13 & 30 & 17.5 \\
\hline
\fontsize{7pt}{8pt}\selectfont
HOUSE-test & 20 & 77 & 1 & 11 & 3.8 \\
\hline
\fontsize{7pt}{8pt}\selectfont
SCOPE-train-5 & 150 & 750 & 5 & 5 & 5.0 \\ 
\fontsize{7pt}{8pt}\selectfont
SCOPE-paraph-5 (NMT-de) & 150 & 3,840 & 13 & 34 & 25.3 \\
\fontsize{7pt}{8pt}\selectfont
SCOPE-paraph-5 (NMT-fr) & 150 & 3,828 & 15 & 33 & 25.52 \\
\fontsize{7pt}{8pt}\selectfont
SCOPE-paraph-5 (LM) & 150 & 4,500 & 30 & 30 & 30.0 \\
\hline
\fontsize{7pt}{8pt}\selectfont
SCOPE-train-10 & 150 & 1,500 & 10 & 10 & 10.0 \\
\fontsize{7pt}{8pt}\selectfont
SCOPE-paraph-10 (NMT-de) & 150 & 7,663 & 33 & 65 & 51.1 \\
\fontsize{7pt}{8pt}\selectfont
SCOPE-paraph-10 (NMT-fr) & 150 & 7,601 & 39 & 64 & 50.7 \\
\fontsize{7pt}{8pt}\selectfont
SCOPE-paraph-10 (LM) & 150 & 9,000 & 60 & 60 & 60.0 \\
\hline
\fontsize{7pt}{8pt}\selectfont
SCOPE-train-50 & 150 & 7,500 & 50 & 50 & 50.0 \\ 
\fontsize{7pt}{8pt}\selectfont
SCOPE-paraph-50 (NMT-de) & 150 & 37,754 & 183 & 314 & 251.7\\
\fontsize{7pt}{8pt}\selectfont
SCOPE-paraph-50 (NMT-fr) & 150 & 37,387 & 206 & 298 & 249.2 \\
\fontsize{7pt}{8pt}\selectfont
SCOPE-paraph-50 (LM) & 150 & 45,000 & 300 & 300 & 300.0\\
\hline
\fontsize{7pt}{8pt}\selectfont
SCOPE-test & 150 & 4,500 & 30 & 30 & 30.0 \\
\hline
\end{tabular}}
\caption{Datasets used in our experiments. HOUSE and SCOPE, respectively, represent our in-house corporate dataset and the in-scope public dataset. NMT-DE represent the machine translation paraphrasing approach with German as the pivot language. FR, EN, and ES represent French, English, and Spanish, respectively. LM represents the paraphrases obtained from using the fine-tuned PEGASUS language model.}
\label{tab:datasets_all}
\end{table*}

\textbf{The SCOPE dataset.} The core training sets are SCOPE-train-5, SCOPE-train-10, and SCOPE-train-50, each one constitutes a baseline for the augmentation task because they have not been artificially augmented. 
We generated the following augmented datasets as described in Sub-sections~\ref{subsec:nmt-system} and \ref{subsec:language-model-system}: 
\begin{itemize}
    \item SCOPE-paraph-5 (NMT-fr): 5 samples per intent version paraphrased with French as the pivot language to generate paraphrases in English.
    \item SCOPE-paraph-5 (NMT-de): 5 samples per intent version paraphrased with German as the pivot language to generate paraphrases in English.
    \item SCOPE-paraph-5 (LM): 5 samples-per-intent version augmented using PEGASUS.
    \item SCOPE-paraph-10 (NMT-fr): 10 samples per intents paraphrased with French as the pivot language to generate paraphrases in English.
    \item SCOPE-paraph-10 (NMT-de): 10 samples per intents paraphrased with German as the pivot language to generate paraphrases in English.
    \item SCOPE-paraph-10 (LM): 10 samples per intents augmented using PEGASUS.
    \item SCOPE-paraph-50 (NMT-fr): 50 samples per intents paraphrased with French as the pivot language to generate paraphrases in English.
    \item SCOPE-paraph-50 (NMT-de): 50 samples per intents paraphrased with German as the pivot language to generate paraphrases in English.
    \item SCOPE-paraph-50 (LM): 50 samples per intents augmented using PEGASUS.
\end{itemize}

 The baselines SCOPE-train-5, SCOPE-train-10, and SCOPE-train-50 as well as all the augmented datasets are in English.

In Table~\ref{tab:datasets_all}, we lay out the HOUSE and SCOPE core and augmented datasets, the paraphrasing volumes for each of the partitions using the NMT and the PEGASUS-based approaches.

\subsection{Dialog System Setup}

To evaluate the potential impact of our augmented datasets on the performance of dialog systems, we created dialog systems using the Rasa framework~\cite{bocklisch2017rasa} and trained the NLU models with the core training datasets. 
The Rasa framework makes use of the DIET Classifier for intent classification~\cite{bunk2020diet}. 
As for the parametric configuration of the framework, we used a Bag-of-Words encoder to obtain sparse representations at both word and character levels. These representations will be used as input by the classifier who will then train it's own dense embeddings. 
All NLU models were trained for 200 epochs. 
For each of our baselines (core HOUSE and the three core SCOPEs), we trained a Rasa-based dialog system using said configuration. 
This yields one HOUSE baseline system and three SCOPE baseline systems (SCOPE-5, -10, and -50). 
Similarly, we trained a dialog system with each of the augmented datasets. 
Again, we must emphasize the fact that we are not trying to reach state-of-the-arts results on our benchmark datasets, but rather demonstrate that artificial data augmentation through paraphrase generation can have enough impact on performance in the early stages of a dialog system to speed-up and reduce the cost of the launch into production.

\begin{table*}[t!]
\centering
{\def\arraystretch{1.2}
\begin{tabular}{|l|c|c|c|c|}
\hline
{\textbf{Systems}} & \textbf{Micro  Score (\%)} & \textbf{Macro  F1 (\%)} & \textbf{Macro  Precision (\%)} & \textbf{Macro  Recall (\%)}\\
\hline
SCOPE-train-5 (baseline) & 54.7 & 52.9 & 54.9 & 54.7 \\
SCOPE-paraph-5 NMT-de & 62.0 & 59.8 & 61.3 & 62.0 \\
SCOPE-paraph-5 NMT-fr & 59.6 & 58.3 & 61.0 & 59.6 \\
SCOPE-paraph-5 LM & \textbf{64.0} & \textbf{62.7} & \textbf{65.4} & \textbf{64.0}\\ \hline

SCOPE-train-10 (baseline) & 69.2 & 68.1 & 69.5 & 69.2   \\
SCOPE-paraph-10 NMT-de & 73.2 & 72.4 & 73.7 & 73.2 \\
SCOPE-paraph-10 NMT-fr & 73.6 & 73.0 & \textbf{75.3} & 73.6 \\
SCOPE-paraph-10 LM & \textbf{74.1} & \textbf{73.1} & 74.6 & \textbf{74.1}\\  
\hline
SCOPE-train-50 (baseline) & 83.8 & 83.4 & 85.1 & 83.8 \\
SCOPE-paraph-50 (NMT-de) & 85.0 &  82.9 & 86.3 & 85.0 \\
SCOPE-paraph-50 (NMT-fr) & 87.1 & 86.9 & \textbf{88.0} & 87.1 \\
SCOPE-paraph-50 (LM) & \textbf{87.4} & \textbf{87.2} & \textbf{88.0} & \textbf{87.4}\\ \hline
HOUSE-train (baseline) & 79.2 & 78.3 & 82.3 & 79.2  \\
HOUSE-paraph (NMT-de) & 84.3 &  81.6 & 81.6 & 83.9 \\
HOUSE-paraph (NMT-en) & \textbf{84.4} & 85.3 & 85.5 & 83.4 \\
HOUSE-paraph (NMT-es) & 83.1 & 82.8 & 83.8 & 81.9 \\
HOUSE-paraph (LM) & \textbf{84.4} & \textbf{85.5} & \textbf{86.7} & \textbf{83.5}\\ \hline
\end{tabular}}
\vspace{0.1in}
\caption{Intent classification results for SCOPE and HOUSE augmented using the various approaches.}
\label{tab:res}
\end{table*}

\section{Evaluation}
\label{sec:eval}
We perform a comparative evaluation between the baselines, i.e. the core un-augmented datasets (HOUSE-train, SCOPE-train-5, SCOPE-train-10, and SCOPE-train-50), and the augmented datasets (those infixed with "paraph" in Table~\ref{tab:datasets_all}).

To evaluate the performance of a dialog system, the effectiveness of intent classification must be measured. 
For that purpose, we use macro-averaged precision, recall, and F1-score. 
We additionally report the micro-averaged score (micro-averaged F1, precision, and recall, which are coequal since each sample is assigned to a single class). 
In our context, precision reflects the ability of the classifier to correctly predict a given intent out of the predictions made, which, in turn translates to the ability of the dialog system to respond with the correct answer. 
A high recall reflects the ability of the intent classifier to ideally predict (realistically, as many as possible) the correct intent for all the test examples associated with that intent. 
The quality of the classification hence depends on that of the paraphrasing component, whether it be NMT- or LM-based.

While most of the artificial data seem to be good paraphrase and even some of the intents were paraphrased perfectly, some paraphrases were spurious, unsurprisingly. 
The noise is often reflected as paraphrases suffering a semantic drift from the original intent or loss of some details or preciseness. 
For instance, with PEGASUS, the sample \emph{"commander des devises"} in French, which translates into \emph{"ordering currencies"} - a quite prevalent financial transaction, had drifted into "I want the dollars", losing its precision, scope and banking-like undertone. 
A noisy example with NMT is \emph{"Spécimen de chèque"} which means \emph{"sample (void) check"}, here the word \emph{chèque} can be translated into its other meanings  \emph{to validate} or \emph{to verify}.

Such noise is undesirable yet unavoidable in an automated augmentation process. 
To quantify the presence of this noise, we asked human expert annotators to determine whether the generated paraphrases for the HOUSE-paraph (LM) dataset were correct or not.
Two annotators with bilingual proficiency in English and French performed the evaluation separately.
They found that 82\% of the paraphrases were correct and reflected the meaning of the original intent while using different syntax and vocabulary.

\section{Results and Discussion} \label{sec:results}

In this Section, we present the evaluation results for both augmentation approaches on the HOUSE and SCOPE datasets. 
Table~\ref{tab:res} reports the overall results in comparison with those obtained with the un-augmented baselines. 
The first observation we make is that the F1-score for the intent classification model has increased by up to 7.8\% (absolute difference) and for all augmented datasets compared to their baselines when evaluated with the test set (HOUSE-test and SCOPE-test). 
Such an increase is observed even for SCOPE-train-50, which carried a noticeably large number of samples per intent pre-augmentation. 
The performance of the various paraphrasing systems themselves seems to be quite close to one another, though the LM-based system yields the best F1 score for most datasets. 
Furthermore, we point out that datasets with fewer training examples per intent seemed to benefit more from paraphrasing than the larger datasets, which already possess more generalization capabilities due to the higher volumes of data. 
This confirms the hypothesis that paraphrasing is more useful in the early stages of dialog system development process, when very few training samples per intent are available. 

Table~\ref{tab:res} shows the scores obtained with the HOUSE dataset, which is the smaller dataset with only 20 intents and very few samples. Again, we can see that paraphrasing always results in a higher micro score. 
This shows that either paraphrasing systems can be used with success on dialog systems in a language other than English.
When comparing the individual systems, the performance levels are again quite close but the LM-based system still outperforms the NMT-based ones by a small margin, despite the added translation step needed for French. However, it is important to mention that the NMT-based approach use lighter models than the LM-based one, which could make it a viable choice for groups or organizations with limited resources.

\section{Conclusions and Future Work} \label{sec:conclusion}

This work explored the benefit of paraphrasing as a data augmentation solution when creating NLU models for dialog systems in a real industrial context. 
We experimented with two paraphrasing approaches: one is NMT-based, and the other leverages large pre-trained language models for text generation tasks. 
We used these paraphrasing methods to augment two datasets: firstly, a publicly available dataset in English, with a large number of intents and partitions of varying sizes; secondly, a corporate dataset in French, with a smaller number of intents. 
In the light of our experiments, we found that the benefit of paraphrasing is undeniable, as the models trained on the augmented sets, across all datasets and paraphrasing systems, outperformed the baseline models (trained with un-augmented data) for the intent classification task. 
This data augmentation method can be used effectively by organizations to create new dialog systems, without having to deal with the huge costs and resource needs associated with manual data generation.

We summarize our findings as follows:

\begin{enumerate}
\setlength\itemsep{1em}
\item  Our proposed augmentation approaches outperformed the un-augmented baselines on all datasets for both French and English. This shows that the quality and diversity of the paraphrases can improve the NLU model generalization capabilities.
\item  Both NMT- and PEGASUS-based approaches improved the performance. Using PEGASUS slightly outperforms NMT as an augmentation component, despite the added translation step at the beginning and at the end of the pipeline in the case of the French dataset.
\item NMT provides a smaller semantic variety than does PEGASUS, but is also safer than PEGASUS as the latter might stray away from the original intent.
\item In NMT-based approaches, no clear difference between the various pivot languages could be observed. French and English, however, seem to pair well as source-pivot languages in both directions.
\item The more intents there are, the more training samples are needed, which comes as no surprise. However, we have seen in our experiments that 10 original samples per intent and their paraphrase augmentations can be enough to reach a performance that allows a satisfying user experience, hence start interacting with real end-users.
\end{enumerate}
\vspace*{0.12in}
We therefore recommend that five to 10 initial samples per intent be handcrafted and phrased diversely to convey various aspects of the intent, then augmentation is to be conducted using one of our proposed approaches.

We also show that our augmentation approaches are low-cost and significantly faster than building a substantial training set manually. 
As for future work, we would like to experiment with our augmentation methods on more mature dialog systems with more samples per intent, based on actual user interactions, as new intents emerge and existing ones evolve. 
Also, user logs often bring a different type of noise into the data, such as misspelling and specialized terms. 
It would thus be interesting to explore the impact of such noise.
Then, it could be interesting to test ways to filter out bad paraphrases during the augmentation pipeline. This could be done, for example, by tuning a threshold on the model confidence score instead of extracting the top n paraphrases.
Finally, it would be interesting to attempt a fusion approach, that is to combine unique paraphrases originating from different and more diverse pivots (e.g. Chinese and French) for higher paraphrasing diversity attributed to language subtleties, and possibly to combine paraphrases from both NM- and LM-based approaches into the training set. 

\newpage

\bibliographystyle{unsrtnat}
\bibliography{quick_starting_bots_arxiv}  

\end{document}